\documentclass[journal]{IEEEtran}
\usepackage{graphicx}
\usepackage{float}
\usepackage{subfigure}

\usepackage{amssymb}
\usepackage{array}
\usepackage{amsmath}
\usepackage{url}
\usepackage{multirow}
\usepackage{booktabs}
\usepackage{threeparttable}
\usepackage{fixltx2e}
\usepackage{float}
\usepackage{booktabs}
\usepackage{graphicx}
\usepackage{diagbox}
\usepackage{mathtools}
\usepackage{amsmath}
\usepackage{bbding}
\usepackage{pifont}
\usepackage{wasysym}
\usepackage{tikz}
\usepackage{calc}
\usepackage{amssymb}
\usepackage{multirow}
\usepackage{rotating}
\usepackage{colortbl}
\usepackage{color}
\usepackage{ragged2e}
\usepackage[colorlinks,linkcolor=red,anchorcolor=blue,citecolor=green]{hyperref}
\usepackage{cite}
\usepackage{gensymb}
\usepackage{makecell}

\def\eg{\emph{e.g.}}
\def\ie{\emph{i.e.}}
\def\etal{{\em et al.}}

\usepackage{amsmath}
\usepackage{amssymb}
\usepackage{multirow}
\usepackage{array}

\definecolor{mygray2}{gray}{.6}
\definecolor{mygray3}{gray}{.3}

\def\ourmethod{HTR}
\def\ourmemory{hybrid memory}

\begin{document}
\title{Temporally Consistent Referring Video Object Segmentation with Hybrid Memory}
\vspace{10mm}
\author{Bo Miao, 
Mohammed Bennamoun,
Yongsheng Gao,
Mubarak Shah, and
Ajmal Mian
\IEEEcompsocitemizethanks{\IEEEcompsocthanksitem Bo Miao, Mohammed Bennamoun, and Ajmal Mian are with the Department of Computer Science and Software Engineering, The University of Western Australia, Perth, Crawley, WA 6009 Australia (e-mail: bomiaobbb@gmail.com, mohammed.bennamoun@uwa.edu.au, ajmal.mian@uwa.edu.au). 
\IEEEcompsocthanksitem Yongsheng Gao is with the School of Engineering, Griffith University, Brisbane, QLD 4111 Australia (e-mail: yongsheng.gao@griffith.edu.au). 
\IEEEcompsocthanksitem Mubarak Shah is with the Center for Research in Computer Vision, University of Central Florida, Orlando, FL 32816 USA (e-mail: shah@crcv.ucf.edu). 
}
\thanks{This research was funded by the Australian Research Council Industrial Transformation Research Hub IH180100002. Professor Ajmal Mian is the recipient of an Australian Research Council Future Fellowship Award (project number FT210100268) funded by the Australian Government.}
}

\markboth{} 
{Shell \MakeLowercase{\textit{et al.}}: Bare Demo of IEEEtran.cls for IEEE Journals}

\maketitle
\begin{abstract}
Referring Video Object Segmentation (R-VOS) methods face challenges in maintaining consistent object segmentation due to temporal context variability and the presence of other visually similar objects. 
We propose an \emph{end-to-end} R-VOS paradigm that explicitly models temporal instance consistency alongside the referring segmentation. 
Specifically, we introduce a novel hybrid memory that facilitates inter-frame collaboration for robust spatio-temporal matching and propagation.
Features of frames with automatically generated high-quality reference masks are propagated to segment the remaining frames based on multi-granularity association to achieve temporally consistent R-VOS.
Furthermore, we propose a new Mask Consistency Score (MCS) metric to evaluate the temporal consistency of video segmentation. 
Extensive experiments demonstrate that our approach enhances temporal consistency by a significant margin, leading to top-ranked performance on popular R-VOS benchmarks, \ie, Ref-YouTube-VOS (67.1\%) and Ref-DAVIS17 (65.6\%).
The code is available at \url{https://github.com/bo-miao/HTR}.
\end{abstract}
\begin{IEEEkeywords}
Referring video object segmentation, temporal consistency, deep learning.
\end{IEEEkeywords}

\section{Introduction}
\IEEEPARstart{R}{eferring} video object segmentation (R-VOS) is a multimodal reasoning task, which aims at segmenting target objects in videos based on linguistic expressions. 
Unlike referring image segmentation~\cite{LAVT,VLT,tcsvt12,tcsvt13}, R-VOS poses a greater challenge as it requires maintaining temporal consistency and coherence throughout the entire video sequence.
In contrast to semi-supervised video object segmentation~\cite{tcsvt1,tcsvt2,tcsvt3,tcsvt4,tcsvt5,XMem,AOT,RAVOS} (Semi-VOS), 
R-VOS adds difficulty by not providing any ground-truth mask for the initial frame as reference.

Early R-VOS methods~\cite{LREVOS,HuA2D} primarily employ convolution networks~\cite{ResNet} for visual encoding and language grounding. 
However, their inherent limitations in capturing long-range dependencies and handling free-form features result in suboptimal performance.
With the advancement in attention mechanism~\cite{Attention,OTS,mingtao,liu2024context}, various attention-based R-VOS approaches have emerged, leading to significant improvements.
Techniques such as YOFO~\cite{YOFO} and SgMg~\cite{sgmg} employ cross-attention to model multimodal representations, enabling flexible and long-range vision-language correspondence. 
Furthermore, ReferFormer~\cite{ReferFormer} and MTTR~\cite{MTTR} introduce end-to-end transformers with object queries to dynamically separate target objects from semantic features.

\begin{figure*}[t!]
\centering
\includegraphics[width=1\textwidth]{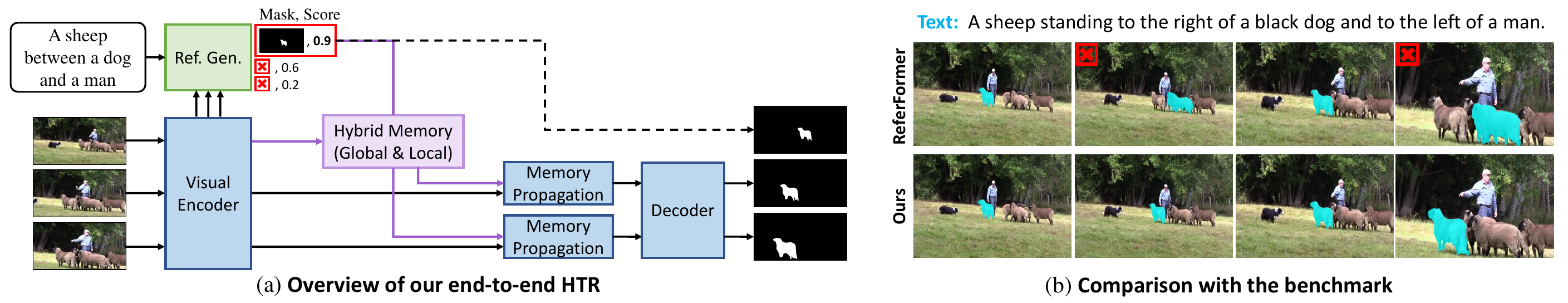}
\caption{(a) Our \ourmethod{} automatically generates reference masks and achieves temporally consistent R-VOS in an end-to-end manner using the robust \ourmemory{}.
(b) The baseline model ReferFormer~\cite{ReferFormer} loses track of the target in some frames (marked by \textbf{\textcolor{red}{$\XBox$}}) whereas ours maintains temporal consistency in segmenting the correct object.}
\label{fig:overview}
\end{figure*}

These transformer-based techniques employ a set of language-guided object queries to attend to multimodal features of different frames, eventually predicting conditional kernels~\cite{ConditionConv} for each frame.
The frame-specific conditional kernels are then used to generate object masks by convolving the corresponding multimodal features.
Despite demonstrating favorable performance, these approaches struggle to maintain consistent segmentation in complex and challenging scenarios.
As illustrated in the first row of Fig.~\ref{fig:overview}(b), the target sheep is accurately segmented in the initial frame, but the presence of other similar-looking sheep and changes in the local and global context, \eg, the disappearance of the dog in the final frames, lead to the failure of the model to consistently segment the correct target.

To mitigate this inconsistency issue, some approaches utilize well-trained Semi-VOS models~\cite{cfbi,STCN} for post-processing,
\eg, ReferFormer~\cite{ReferFormer} and PMINet~\cite{PMINet} use CFBI~\cite{cfbi} as a post-processing step to refine the masks. 
Such a na\"ive strategy requires a complex pipeline with additional backbones~\cite{ResNet,deeplab} and training data~\cite{DUTS,ECSSD,HRSOD,BIG,FSS1000,MIVOS}, making R-VOS cumbersome and impractical.
Moreover, this strategy does not yield satisfactory improvements in performance as we show in Table~\ref{tab:ablationexternalvos}. 
An ideal solution would be an end-to-end paradigm that seamlessly integrates temporal consistency into R-VOS.

We introduce \textbf{\underline{H}}ybrid memory for \textbf{\underline{T}}emporally consistent \textbf{\underline{R}}eferring video object segmentation (\textbf{HTR}).
Instead of relying on well-trained Semi-VOS models, \ourmethod{} integrates inter-frame association and referring segmentation into an end-to-end paradigm. 
This not only eliminates the need for additional backbones and training data, but also results in a significant gain in accuracy.
In contrast to prior temporal propagation methods~\cite{STCN,cfbi}, \ourmethod{} includes a new \ourmemory{} that stores not only fine-grained local context but also lightweight global representations for explicit propagation. 
This memory exhibits higher robustness for feature propagation in the presence of noise that stems from imperfect automatically generated reference masks (as no ground-truth mask is provided for R-VOS during inference) and correlation between similar-looking objects.
By leveraging inter-frame collaboration, highly confident segmentation in the \ourmemory{} is effectively propagated to the remaining frames, leading to improved temporal consistency and mask quality.

To evaluate the proposed \ourmethod{}, we conduct a series of experiments on popular R-VOS benchmarks, Ref-YouTube-VOS~\cite{URVOS}, Ref-DAVIS17~\cite{LREVOS}, A2D-Sentences~\cite{HuA2D}, JHMDB-Sentences~\cite{JHMDB}.
Without bells and whistles, \eg, test-time augmentations and model ensembles, \ourmethod{} achieves top-ranked performance: 67.1\% on Ref-YouTube-VOS, 65.6\% on Ref-DAVIS17, 59.0\% on A2D-Sentences, and 44.9\% on JHMDB-Sentences.
Notably, \ourmethod{} outperforms the baseline model ReferFormer~\cite{ReferFormer} by up to 5.1\% points. 
\textcolor{black}{Furthermore, we introduce a new metric, Mask Consistency Score (MCS), to evaluate the temporal consistency of video segmentation at different thresholds $\tau$ of individual frame segmentation accuracy.}
\ourmethod{} shows significantly improved temporal consistency, with relative improvements of up to 53\% in terms of MCS@0.9 ($\tau = 0.9$). Our main contributions are summarized below:
\begin{itemize}
\item We propose \ourmethod{}, a highly effective and \emph{end-to-end} paradigm that explicitly models temporal instance consistency alongside referring segmentation. 
\ourmethod{} achieves top-ranked performance on the benchmark datasets while also having an efficient run-time speed.
\item We develop a \ourmemory{}, in conjunction with inter-frame collaboration, to facilitate robust spatio-temporal propagation.
The proposed memory significantly improves temporal consistency and mask quality.
\item We propose a new Mask Consistency Score (MCS) metric to evaluate the temporal consistency of video segmentation.
This will encourage progress in R-VOS by benchmarking future advancements in the temporal consistency of video segmentation.
\end{itemize}

\section{Related Work}

\subsection{Semi-supervised Video Object Segmentation} 
Semi-VOS aims to segment target objects throughout the entire video, given the ground-truth mask (of the first frame)~\cite{tcsvt6,tcsvt7,tcsvt8,tcsvt9}.
Existing techniques mainly fall into three paradigms: finetuning-based, tracking-based, and matching-based.
Finetuning-based methods dynamically update networks during test time using the first~\cite{OSVOS,OSVOS_S}, selected~\cite{E_OSVOS}, or synthetic~\cite{PReMVOS} frames from each video to make models object-specific.
Despite their promising results,
the slow test-time adaptation makes them inefficient.
Tracking-based approaches~\cite{cfbi,MaskTrack,RGMP,OSMN,SAT,FAVOS,AGSS,MHPVOS,DTN,FEELVOS,Ranet} avoid online learning and perform frame-to-frame propagation using masks of neighboring and annotated frames.
However, they often lack sufficient long-term context, leading to the loss of targets after occlusions. 

To address the context limitations, matching-based methods~\cite{KMN,MAMP,Capsulevos,TAVOS,RPMVOS,LPTC,AFB_URR,EMAMP,RAVOS} construct memory banks with features from past frames and use attention mechanisms for temporal propagation.
STM~\cite{STM} and STCN~\cite{STCN} build memory banks to store mask and visual features at fixed intervals. SST~\cite{SST} uses transformer blocks for affinity computation. AOT~\cite{AOT} and its subsequent works~\cite{DeAOT} establish hierarchical propagation to process multiple objects in parallel. 
XMem~\cite{XMem} introduces complementary memories inspired by the Atkinson-Shiffrin model~\cite{Atkinson-Shiffrin} to efficiently handle long videos.
\textcolor{black}{In this work, we explore temporally consistent R-VOS, a task more challenging than Semi-VOS due to multimodal reasoning and the lack of ground-truth reference mask during inference, and propose an end-to-end paradigm based on a \ourmemory{}. Unlike STM~\cite{STM} and STCN~\cite{STCN}, our \ourmemory{} employs both local and global contexts for robust propagation of imperfect reference masks. In contrast to CFBI~\cite{cfbi} and PMNet~\cite{PMNet}, the global tokens in our memory are constructed in one-stage and perform explicit node-object matching.}

\begin{figure*}[t]
\centering
\includegraphics[width=1\textwidth]{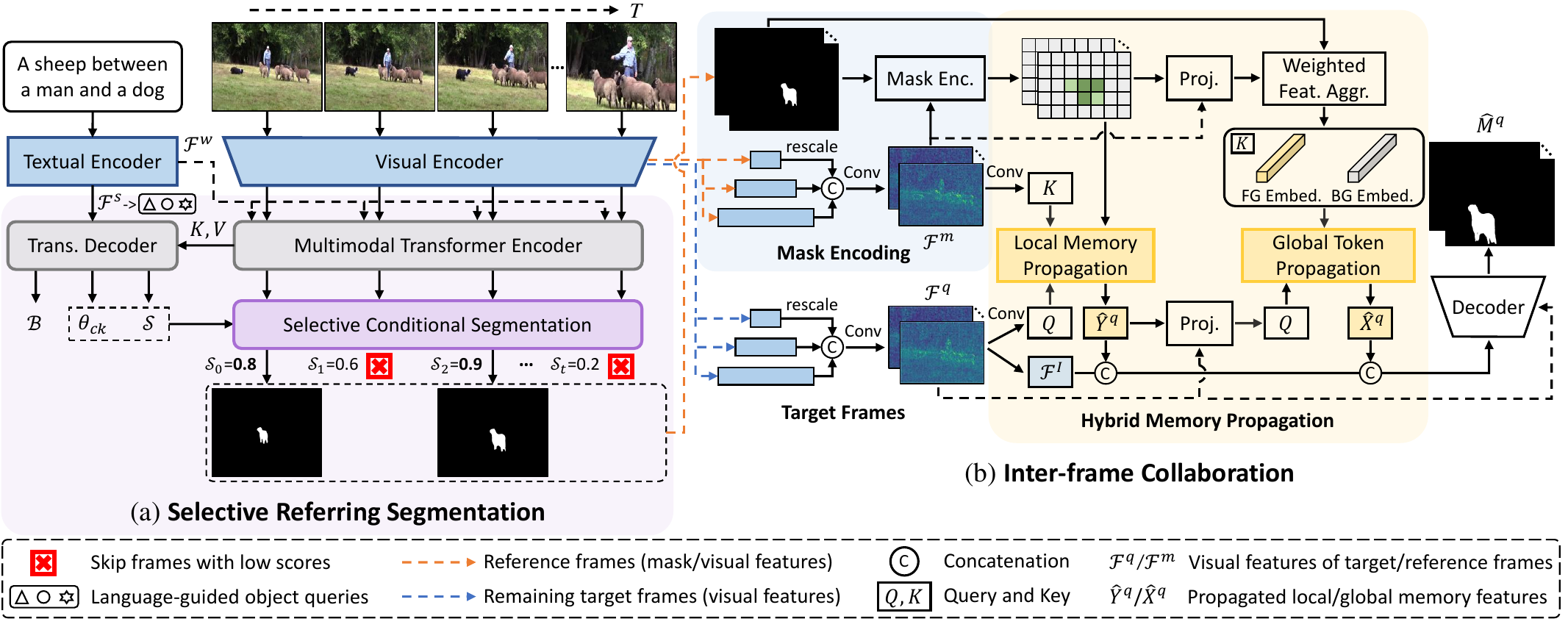}
\caption{Detailed architecture of \ourmethod{}. 
(a) Selective referring process predicts the score $\mathcal{S}$ and conditional kernels $\theta_{ck}$ for each frame to selectively segments frames with high scores.
Masks and visual features of these selected reference frames, and only the visual features of the remaining target frames are passed to (b) Inter-frame collaboration module. 
This module encodes reference frames to construct \ourmemory{} and aggregates memory features based on node-node (\emph{local memory}) and node-object (\emph{global token}) affinity to segment each pixel node in target frames.
$\mathcal{F}^{s}$/$\mathcal{F}^{w}$: sentence/word features.
}\label{fig:framework}
\end{figure*}

\subsection{Referring Video Object Segmentation} 
R-VOS methods conduct vision-and-language interactions to segment the referred object throughout the video~\cite{ReferFormer,sgmg,TCD}.
LRE~\cite{LREVOS} performs referred object localization and regional segmentation.
URVOS~\cite{URVOS} combines vision-language and temporal features for iterative segmentation. 
UMVOS~\cite{tcsvt14} adopts reinforcement learning to optimize multi-modality filter networks.
MMTBVS~\cite{MMTBVS} uses motion clues to enhance multimodal representations.
MLRL~\cite{MLRL} extracts video-level, frame-level, and object-level features for comprehensive representations.
Clawcranenet~\cite{Clawcranenet} and CMPC~\cite{CMPC} mimic human behavior to progressively execute discriminative vision-language fusion.
Similarly, HINet~\cite{HINet}, YOFO~\cite{YOFO}, and LBDT~\cite{LBDT} leverage multi-layer features for hierarchical cross-modal fusion.

Among these, transformer-based approaches have gained significant attention for their end-to-end architecture and impressive performance. 
MTTR~\cite{MTTR} introduces spatio-temporal transformer~\cite{VideoSwin} for visual encoding and conditional kernels~\cite{ConditionConv} for dynamic object segmentation.
ReferFormer~\cite{ReferFormer} employs language as queries to segment referred objects and includes a detection loss to facilitate object localization learning. 
MANet~\cite{MANet} compresses video data and focuses on keyframe segmentation to speed up R-VOS.
UNINEXT~\cite{UNINEXT} introduces a unified model capable of training with data from various tasks.
R$^{2}$VOS~\cite{R2VOS} enhances multimodal alignment through text reconstruction. 
OnlineRefer~\cite{onlinerefer} proposes an online model with explicit query propagation.
HTML~\cite{HTML} performs multimodal interactions across various temporal scales.
SgMg~\cite{sgmg} proposes a segment-and-optimize paradigm to solve the feature drift issue.
\textcolor{black}{DsHmp~\cite{DsHmp} decouples video-level expression understanding into static and motion perception to enhance temporal comprehension.}
This work proposes an end-to-end paradigm that seamlessly incorporates temporal consistency modeling to achieve consistent and accurate R-VOS.

\section{Method}

\emph{How can we devise an effective paradigm that accurately and \textbf{consistently} segments referred objects in videos?}
We answer this by proposing an end-to-end paradigm called HTR, which employs a novel memory module to propagate features of frames with high-quality masks to the remaining frames for achieving temporally consistent and accurate R-VOS, as shown in Fig.~\ref{fig:framework}.
\ourmethod{} performs selective segmentation based on the confidence measure to generate high-quality reference masks for constructing the \ourmemory{}.
The remaining frames with low scores are segmented by aggregating features from the memory based on robust inter-frame affinity.
Compared to prior R-VOS methods~\cite{ReferFormer,MTTR,MANet,sgmg,CITD}, \ourmethod{} can consistently segment the correct target across frames in an end-to-end manner, resulting in significant improvements in temporal consistency and segmentation quality.

\subsection{Architecture}
Figure~\ref{fig:framework} shows the architecture of our \ourmethod{}, which consists of three components: feature encoding, selective referring segmentation, and inter-frame collaboration.
The feature encoding stage extracts visual and textual features from the input frames and language expressions.
The selective referring segmentation stage is built upon ReferFormer~\cite{ReferFormer}, which follows the standard transformer-based video segmentation framework~\cite{VISTR,MTTR}.
This stage employs language-guided object queries for conditional segmentation on frames with high scores.
Notably, \ourmethod{} is a general paradigm and is also compatible with other networks such as the recent SgMg~\cite{sgmg}.
The inter-frame collaboration incorporates a new \ourmemory{} to perform robust inter-frame association and propagation for temporally consistent R-VOS.

\subsection{Visual and Textual Encoders}
We adopt Swin Transformer~\cite{Swin} or VideoSwin Transformer~\cite{VideoSwin} as the visual encoder to extract visual features $\mathcal{F}^{v}$ and RoBERTa~\cite{Roberta} as the textual encoder for generating word-level features $\mathcal{F}^{w} \in{\mathbb{R} ^{N \times C}}$ and sentence embeddings $\mathcal{F}^{s} \in{\mathbb{R} ^{1 \times C}}$, owing to their strong representation ability. 
The extracted visual and linguistic features are projected to a lower dimension of 256 to facilitate efficient multimodal interactions.
For each frame, the visual encoder produces a feature pyramid with spatial strides of \{8,16,32\}.

\subsection{Selective Referring Segmentation}
\label{sec:selectiveseg}
\subsubsection{Multimodal Transformer}
\textcolor{black}{To generate object-related multimodal representation $\mathcal{F}^{vl}_{l}$, cross-attention is employed to enhance each layer of the visual feature pyramid $\mathcal{F}^{v}$ by incorporating linguistic information $\mathcal{F}^{w}$,}
\begin{equation} \label{equ:vlfusion}
\begin{aligned}
\mathcal{F}^{vl}_{l} = \mathcal{F}^{v}_{l} \odot	{\rm Attn}(\mathcal{F}^{v}_{l}W^{Q}_{l}, \mathcal{F}^{w}W^{K}_{l}, \mathcal{F}^{w}W^{V}_{l})
\end{aligned}
\end{equation}
where $\odot$ is the Hadamard product and $W$ represents learnable projection matrices. ${\rm Attn}(Q,K,V)$ is the standard attention block~\cite{Attention} with query, key, and value.

The multimodal transformer integrates information from the vision-language feature pyramid $\mathcal{F}^{vl}$ to effectively handle objects of varying sizes and scales. In this work, we adopt the memory-efficient Deformable Transformer~\cite{DeformableTransformer} to encode vision-language features $\mathcal{F}^{vl}$ and predict object embeddings $\mathcal{E}$, similar to~\cite{ReferFormer,mask2former,MDETR}. 
\textcolor{black}{The transformer encoder takes $\mathcal{F}^{vl}$ as input and outputs a feature pyramid of the same size.
The transformer decoder then uses a set of language-guided object queries $\mathcal{Q}$, generated by combining sentence features $\mathcal{F}^{s}$ and learnable embeddings, and cross-attends them with encoded vision-language features $\mathcal{F}^{vl}$ to predict frame-wise object embeddings $\mathcal{E}$,}
\begin{equation} \label{equ:transdecode}
\begin{aligned}
\mathcal{E} = \mathcal{Q} +	{\rm Attn}(\mathcal{Q}W^{Q}, \mathcal{F}^{vl}W^{K}, \mathcal{F}^{vl}W^{V})
\end{aligned}
\end{equation}
Each object embedding uses fully connected layers to predict conditional kernels $\theta_{ck}$, a score $\mathcal{S}$, and a bounding box $\mathcal{B}$ for an individual frame.
\textcolor{black}{The kernels $\theta_{ck}$ and score $\mathcal{S}$ are employed for \emph{selective conditional segmentation},} while the bounding box $\mathcal{B}$ facilitates object localization learning following~\cite{ReferFormer}.

\subsubsection{Selective Conditional Segmentation}
Considering the temporal context variability, language-guided object queries and features may fail to consistently identify the correct targets, leading to temporally inconsistent and incorrect segmentation outcomes.
\textcolor{black}{To tackle this, we selectively apply conditional kernels $\theta_{ck}$ for referring segmentation on corresponding frames predicted to have high segmentation quality score $\mathcal{S}$.}
Specifically, we supervise the score $\mathcal{S}$ during training based on the segmentation loss.
The ground-truth score $\hat{\mathcal{S}}$ for the optimal object query with the minimum loss $\mathcal{L}$ is set to one, while all others are set to zero:
\begin{equation} 
\begin{aligned}
{\hat{\mathcal{S}}}_{i} = 
\begin{cases} 
1, & \text{if } {\mathcal{L}}_{i} = \min({\mathcal{L}}_{1}, {\mathcal{L}}_{2}, ..., {\mathcal{L}}_{N}) \\
0, & \text{otherwise}
\end{cases}
\end{aligned}
\end{equation}
Consequently, higher scores $\mathcal{S}$ become indicative of superior segmentation quality.
During inference, we rank the conditional kernels $\theta_{ck}$ based on their scores $\mathcal{S}$ and selectively use the top-ranked kernels to segment their corresponding frames.
For conditional segmentation, we use a cross-modal FPN decoder~\cite{ReferFormer} to upsample the encoded vision-language features $\mathcal{F}^{vl}$ of the selected frames.
The corresponding kernels $\theta_{ck}$ are then employed as weights of point-wise convolutions applied to the decoded $\mathcal{F}^{vl}$ to predict object masks $\hat{M}$.
This selective segmentation stage efficiently produces reliable reference masks, facilitating subsequent inter-frame collaboration.

\subsection{Inter-frame Collaboration}
\label{sec:IFC}
In our approach, frames with high scores are selectively segmented, and their features are propagated to segment the remaining target frames using the \ourmemory{}. 
As depicted in Fig.~\ref{fig:framework}(b), the target frames aggregate memory features ($\hat{X}^{q}$ and $\hat{Y}^{q}$) from the selected reference frames based on their inter-frame affinity. 
These aggregated features are then concatenated with the visual features of the target frames and passed to a convolutional decoder~\cite{FPN} to predict object masks.

\subsubsection{Mask Encoding}
\ourmethod{} reuses multi-scale visual features for feature matching and enhancement. 
Unlike conventional memory networks~\cite{STM,STCN,XMem} that use extra backbones for object mask encoding, \ourmethod{} adopts an efficient fusion-based strategy. 
It directly projects 16$\times$16 mask grids to hidden embeddings using a learnable transformation matrix and enhances these embeddings with visual features through two residual blocks~\cite{ResNet} and a channel-spatial attention block~\cite{CBAM}.

\subsection{Hybrid Memory}
\label{sec:GLM}
To effectively propagate spatio-temporal features in the presence of noise, which stems from imperfect reference masks and correlation between similar-looking objects, we introduce a \ourmemory{}. As shown in Fig.~\ref{fig:framework}(b), our memory comprises two complementary elements, local memory and global tokens, that collaboratively compute multi-granularity affinity for effective propagation.

\subsubsection{Local Memory} 
The local memory captures pixel-level context for fine-grained feature matching and propagation, \textcolor{black}{preserving object structure and appearance details.}
Let $\mathcal{F}^{q} \in{\mathbb{R} ^{HW \times C}}$ denotes the visual features of a target frame, and $\mathcal{F}^{m},Y^{m} \in{\mathbb{R} ^{THW \times C}}$ denote the memorized visual and mask features of the reference frames, respectively, where $T$, $HW$, $C$ refer to the temporal, spatial, and channel dimensions. The propagation of local memory can be formulated as  
\begin{equation} \label{equ:localmem_sim}
\begin{aligned}
{\rm Sim}(Q,K)_{ij} = -\left\lVert Q_{i} - K_{j} \right\rVert_2^2
 \end{aligned}
\end{equation}
\begin{equation} \label{equ:localmem_prop}
\begin{aligned}
\hat{Y}^{q} = \sigma({\rm Sim}(\mathcal{F}^{q}W^{K},\mathcal{F}^{m}W^{K})) Y^{m}
\end{aligned}
\end{equation}
where $W^{K} \in{\mathbb{R} ^{C \times 64}}$ is a trainable projection matrix used to extract the query and key for feature matching, $\sigma$ denotes the Softmax operation applied along the spatio-temporal dimension of the local memory, and $Sim(Q,K)$ represents the L2 similarity~\cite{STCN} used to compute the affinity.

\subsubsection{Global Tokens} 
\begin{figure}[t!]
\centering
\includegraphics[width=\columnwidth]{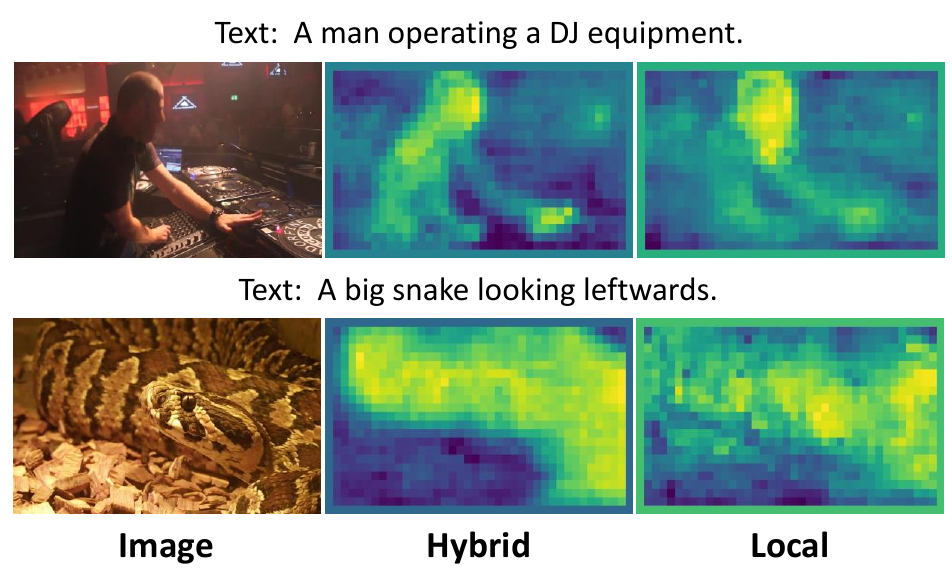}
\caption{Visualization of the propagated features using our \ourmemory{} and a standard local memory~\cite{STCN}. Our memory demonstrates robust feature propagation in challenging scenarios.}\label{fig:memvisual} 
\vspace{-1mm}
\end{figure}

The global tokens extract robust and lightweight representations for foreground and background, each condensed into a single token, providing crucial global spatio-temporal context.
This facilitates target localization and alleviates error propagation.
Specifically, based on the foreground and background probabilities ($M_{i=1}^{m}$ and $M_{i=0}^{m}$) obtained from selective referring segmentation, \ourmethod{} performs weighted feature aggregation to extract foreground and background global representations.
The aggregation process $\Phi_{aggr}(\mathcal{F}; M_{i})$ is achieved through a weighted average of vision-mask joint representations from all reference frames,
\begin{equation}
\small
\label{equ:aggregate}
\begin{aligned}
\Phi_{aggr}(\mathcal{F}; M_{i}^{m}) = \frac{\sum_{j=1}^{THW}{(\mu(M_{i,j}^{m}-\tau)\odot M_{i,j}^{m}\odot \mathcal{F}_{j}})}{\sum_{j=1}^{THW}{(\mu(M_{i,j}^{m}-\tau)\odot M_{i,j}^{m})}}
\end{aligned}
\end{equation}
where $\odot$ denotes Hadamard product, $\tau=0.5$ is the threshold, and $\mu(x)$ represents the Heaviside step function to extract foreground and background binary masks,
\begin{equation} \label{equ:H1}
\mu(x) = \begin{cases}
1, & \text{if } x > 0, \\
0, & \text{if } x \leq 0.
\end{cases}
\end{equation}
Through our weighted feature aggregation, noisy features with low scores contribute less to the global embeddings, resulting in robust foreground and background representations.
The propagation of global tokens can be formulated as 
\begin{equation}
\label{equ:globalmem_sim}
\begin{aligned}
\hat{X}^{q} = (\hat{Y}^{q} \mathbin\Vert \mathcal{F}^{q})W^{J} \Phi_{aggr}((Y^{m} \mathbin\Vert \mathcal{F}^{m})W^{J}; M^m)^{\mathsf{T}}
\end{aligned}
\end{equation}
where $\mathbin\Vert$ denotes the concatenation of visual and mask features and $W^{J} \in{\mathbb{R} ^{2C \times 64}}$ is a trainable projection matrix to extract vision-mask joint representations. 
Note that we only compute the efficient node-object affinity using the global tokens to enhance the propagated fine-grained local memory features $\hat{Y}^{q}$.
After \ourmemory{} propagation, the propagated features ($\hat{X}^{q}$ and $\hat{Y}^{q}$) are concatenated with the visual features of the target frame. These features are then decoded using a common convolutional decoder~\cite{FPN} to predict the object mask $\hat{M}^{q}$.
By leveraging our \ourmemory{}, \ourmethod{} effectively handles spatio-temporal propagation (see Fig.~\ref{fig:memvisual}) for temporally consistent R-VOS.

\subsection{Mask Consistency Score}

Temporal consistency in video segmentation is crucial to maintain accurate and consistent object masks throughout the video, despite changes in appearance or context.
Inconsistent segmentation can cause flickering and loss of tracked targets.
However, existing R-VOS models primarily use frame-wise accuracy metrics for evaluation.
To the best of our knowledge, there are currently no dedicated metrics designed to evaluate the temporal consistency of R-VOS models.

To fill this gap, we propose a new metric, Mask Consistency Score (MCS), to evaluate the temporal consistency of video segmentation at specified thresholds.
The segmentation of a video is considered consistent if the accuracy (Jaccard index) for \textbf{\emph{all}} frames exceeds the threshold $\tau$.
Otherwise, the segmentation is deemed inconsistent due to the loss of tracking targets in one or more frames.
The MCS metric is defined as the ratio of videos with consistent object segmentation.
\textcolor{black}{Different from mTC~\cite{varghese2020unsupervised} for semantic segmentation, our metric does not require optical flow and imposes a strong penalty for the loss of targets (temporal inconsistency).}
Let $\mathcal{J}_{x,y}$ denotes the Jaccard index of the segmentation for frame $y$ in video $x$, the MCS at threshold $\tau$ is formulated as 
\begin{equation}
\label{equ:MCS}
\begin{aligned}
{\rm MCS}\text{@}\tau = \frac{1}{N}\sum_{x=1}^{N} \prod_{y=1}^{T} \mu(\mathcal{J}_{x,y} - \tau)
\end{aligned}
\end{equation}
where $N$ denotes the number of videos, $T$ represents the number of frames per video, and $\mu(x)$ is the Heaviside step function mentioned in Eq.~\ref{equ:H1}.
\textcolor{black}{In this work, we adopt thresholds of 0.1, 0.5, and 0.9 for evaluation.
$\tau$=0.1 ensures at least some level of basic consistency (segmentation not completely off-target); $\tau$=0.5 ensures minimum acceptable temporal consistency for tracking; $\tau$=0.9 ensures high-quality results satisfying most real-world applications.}

\subsection{Training Loss Functions}
\label{sec:Loss}
 
The loss function of \ourmethod{} consists of two parts:
\begin{equation} \label{equ:losstrain}
\begin{aligned}
\mathcal{L}_{train} = \mathcal{L}_{refer} +
\mathcal{L}_{prop}
\end{aligned}
\end{equation}
where $\mathcal{L}_{refer}$ and $\mathcal{L}_{prop}$ are losses for selective referring segmentation and inter-frame collaboration.

\subsubsection{Loss for Selective Referring Segmentation} 
Selective segmentation predicts object masks $\mathcal{M}$, bounding boxes $\mathcal{B}$, and scores $\mathcal{S}$ for each video. 
Given a set of predictions $y = \{\forall y_{i}, i \in [1, ..., N] \}$ obtained from $N$=5 object queries, where $y_{i} = {\{\mathcal{M}}_{i,j}, \mathcal{B}_{i,j}, \mathcal{S}_{i,j}\}_{j=1}^{T}$,
we employ the Hungarian algorithm~\cite{Hungarian} to find the optimal assignment with the highest similarity to the ground truth for training. The selective referring segmentation loss can be represented as
\begin{equation} \label{equ:lossreferr}
\begin{aligned}
\mathcal{L}_{refer} = \sum_{i=1}^{N} \lambda_{\mathcal{S}}\mathcal{L}_{\mathcal{S}_{i}} 
 + \boldsymbol{1}_{y_{i}\neq \oslash} [\lambda_{\mathcal{M}}\mathcal{L}_{\mathcal{M}_{i}} +
\lambda_{\mathcal{B}}\mathcal{L}_{\mathcal{B}_{i}}]
\end{aligned}
\end{equation}
where $\mathcal{L}$ and $\lambda_{}$ denote the loss term and weight, $\boldsymbol{1}_{y_{i}\neq \oslash}$ ensures that only the optimal query has mask and bounding box loss.
Following~\cite{ReferFormer,DeformableTransformer,VISTR}, we use dice~\cite{DICE} and focal~\cite{FOCAL} loss for $\mathcal{L}_{\mathcal{M}}$, focal~\cite{FOCAL} loss for $\mathcal{L}_{\mathcal{S}}$, and GIoU~\cite{GIOU} and L1 loss for $\mathcal{L}_{\mathcal{B}}$.

\subsubsection{Loss for Inter-frame Collaboration} 
The masks predicted by inter-frame collaboration are supervised using bootstrapped cross entropy loss and dice loss with equal weighting, following~\cite{STCN,AOT,MIVOS}.

\begin{table*}[t!]
\setlength{\tabcolsep}{8pt}
\centering
\caption{Quantitative comparison on Ref-YouTube-VOS and Ref-DAVIS17. $^*$: pre-trained with RefCOCO/g/+ images. \textcolor{black}{$^{\dagger}$: uses 20 queries and trains for 30 epochs.}
S-\ourmethod{}: integrates our hybrid memory with the recent work SgMg~\cite{sgmg} for comparison.
Our models achieve top-ranked performance in terms of accuracy (\( \mathcal{J} \)\&\( \mathcal{F} \)) and temporally consistency (MCS@$\tau$)
} \label{tab:mainresults}
\begin{tabular}{l | l | c c c c c c| c c c}
\toprule[1.5pt] 
\multirow{2}{*}{} & \multirow{2}{*}{} & \multicolumn{6}{c |}{\textbf{Ref-YouTube-VOS}} & \multicolumn{3}{c}{\textbf{Ref-DAVIS17}} \\ 
\midrule 
Method & Backbone & \( \mathcal{J} \)\&\( \mathcal{F} \) & \( \mathcal{J} \) & \( \mathcal{F} \) & MCS@0.1 & MCS@0.5 & MCS@0.9
&  \( \mathcal{J} \)\&\( \mathcal{F} \) & \( \mathcal{J} \) & \( \mathcal{F} \)  \\
\midrule 
CMSA~\cite{CMSA} & ResNet-50 & 36.4 & 34.8 & 38.1 & - & - & - & 40.2 & 36.9 & 43.5 \\ 
URVOS~\cite{URVOS} & ResNet-50 & 47.2 & 45.3 & 49.2 & - & - & - &  51.5 & 47.3 & 56.0 \\ 
CMPC-V~\cite{CMPC} & I3D & 47.5 & 45.6 & 49.3 & - & - & - & - & - & - \\ 
PMINet~\cite{PMINet} & ResNeSt-101 & 53.0 & 51.5 & 54.5 & - & - & -  & - & - & - \\ 
YOFO~\cite{YOFO} & ResNet-50 & 48.6 & 47.5 & 49.7 & - & - & - & 53.3 & 48.8 & 57.8\\ 
LBDT~\cite{LBDT} & ResNet-50 & 49.4 & 48.2 & 50.6 & - & - & -  & 54.3 & - & - \\ 
MLRL~\cite{MLRL} & ResNet-50 & 49.7 & 48.4 & 51.0 & - & - & -  & 52.8 & 50.0 & 55.4\\  

UMVOS~\cite{tcsvt14} & - & 52.8 & 49.3 &  56.3 & - & - & - & 56.2 & 52.5 & 59.8 \\
MTTR~\cite{MTTR} & VideoSwin-T & 55.3 & 54.0 & 56.6 & 50.6 & 39.0 & 7.1 & - & - & - \\
MANet~\cite{MANet} & VideoSwin-T & 55.6 & 54.8 & 56.5 & - & - & - & - & - & - \\  

ReferFormer~\cite{ReferFormer}  & VideoSwin-T & 56.0 & 54.8 & 57.3 & 50.0 & 38.9 & 5.8 & - & - & -  \\  
R$^{2}$VOS~\cite{R2VOS} & VideoSwin-T & 57.1 & 55.9 & 58.2 & - & - & - & - & - & - \\
SgMg~\cite{sgmg}  & VideoSwin-T & 58.9 & 57.7 & 60.0 & 50.7 & 39.0 & 7.3 & 56.7 & 53.3 & 60.0  \\  
\textcolor{black}{SOC$^{\dagger}$~\cite{soc}} & \textcolor{black}{VideoSwin-T} & \textcolor{black}{59.2} &\textcolor{black}{57.8} &\textcolor{black}{60.5} &\textcolor{black}{52.9}&\textcolor{black}{42.2}&\textcolor{black}{8.0}&\textcolor{black}{\textbf{59.0}} &\textcolor{black}{55.4} &\textcolor{black}{\textbf{62.6}} \\ 
\rowcolor[gray]{0.9} 
\textbf{\ourmethod{} (Ours)} & VideoSwin-T & \textbf{59.8} & \textbf{58.3}  & \textbf{61.3} & \textbf{60.4} & \textbf{47.8} & \textbf{8.2} & 57.2 & 53.8 & 60.6 \\ 

\rowcolor[gray]{0.9} 
\textbf{\textcolor{black}{S-\ourmethod{} (Ours)}} & VideoSwin-T & \textbf{61.5} & \textbf{60.0}  & \textbf{63.0} & \textbf{62.2} & \textbf{50.5} & \textbf{8.5} & 58.5 & \textbf{55.9} & 61.0 \\   

\midrule
\multicolumn{8}{l}{\emph{Additionally pre-trained with RefCOCO/g/+ static images}}\\
\midrule

ReferFormer$^*$~\cite{ReferFormer} & VideoSwin-T & 59.4 & 58.0 & 60.9 & 50.5 & 40.8 & 5.9 & 59.5 & 56.3 & 62.7  \\  
HTML$^*$~\cite{HTML} & VideoSwin-T & 61.2 & 59.5 & 63.0 & - & - & - & - & - & - \\  
R$^{2}$VOS$^*$~\cite{R2VOS} & VideoSwin-T & 61.3 & 59.6 & 63.1 & - & - & - & - & - & - \\
\textcolor{black}{SOC$^{\dagger}$$^*$~\cite{soc}} & \textcolor{black}{VideoSwin-T} & \textcolor{black}{62.4} & \textcolor{black}{61.1}& \textcolor{black}{63.7}& \textcolor{black}{57.6}&\textcolor{black}{46.9}&\textcolor{black}{8.0}& \textcolor{black}{\textbf{63.5}}& \textcolor{black}{\textbf{60.2}}& \textcolor{black}{66.7} \\ 
\rowcolor[gray]{0.9} 
\textbf{\ourmethod{}$^*$ (Ours)} & VideoSwin-T  & \textbf{64.2} & \textbf{62.7} & \textbf{65.7} & \textbf{63.3} & \textbf{53.8} & \textbf{10.9} & 63.2 & 59.4 & \textbf{67.0} \\ 

\midrule
ReferFormer$^*$~\cite{ReferFormer} & VideoSwin-B & 62.9 & 61.3 & 64.6 & 54.8 & 46.3 & 8.0 & 61.1 & 58.1 & 64.1 \\   
OnlineRefer$^*$~\cite{onlinerefer} & VideoSwin-B & 62.9 & 61.0 & 64.7 & 54.4 & 43.2 & 7.1 & 62.4 & 59.1 & 65.6 \\ 
HTML$^*$~\cite{HTML} & VideoSwin-B & 63.4 & 61.5 & 65.2 & - & - & - & 62.1 & 59.2 & 65.1 \\  
VLT$^*$~\cite{VLT} & VideoSwin-B & 63.8 & 61.9 & 65.6 & - & - & - & 61.6 & 58.9 & 64.3 \\
SgMg$^*$~\cite{sgmg} & VideoSwin-B & 65.7 & 63.9 & 67.4 & 55.7 & 48.4 & 11.4 & 63.3 & 60.6 & 66.0  \\ 
SOC$^{\dagger}$$^*$~\cite{soc} & VideoSwin-B & 66.0 & 64.1 & 67.9 & 58.6 & 49.5 & 11.9 & 64.2 &  61.0 &  67.4 \\

\rowcolor[gray]{0.9} 
\textcolor{black}{\textbf{\ourmethod{}$^*$ (Ours)}} & VideoSwin-B  & \textbf{66.7} & \textbf{64.9} & \textbf{68.6} & \textbf{64.8} & \textbf{56.7} & \textbf{12.2} & 63.5 & 60.5  & 66.6  \\ 
\rowcolor[gray]{0.9} 
\textcolor{black}{\textbf{S-\ourmethod{}$^*$ (Ours)}} & VideoSwin-B & \textbf{67.5} & \textbf{65.6} & \textbf{69.4} & \textbf{65.0} & \textbf{56.2} & \textbf{12.8} & \textbf{64.8} & \textbf{61.8} & \textbf{67.9}  \\

\midrule

ReferFormer$^*$~\cite{ReferFormer} & Swin-L & 62.4 &  60.8 & 64.0 & 53.2 & 44.7 & 7.6 & 60.5 &  57.6 & 63.4  \\   
OnlineRefer$^*$~\cite{onlinerefer} & Swin-L & 63.5 & 61.6 & 65.5 & 54.0 & 44.8 & 7.7 & 64.8 & 61.6 & 67.7  \\  
DMFormer$^*$~\cite{tcsvt15} & Swin-L & 64.9 &  63.4 & 66.5 & - & - & - & 62.3 & 59.5 & 65.1 \\
\rowcolor[gray]{0.9}
\textbf{\ourmethod{}$^*$ (Ours)} & Swin-L  & \textbf{67.1} & \textbf{65.3} & \textbf{68.9} & \textbf{64.0} & \textbf{54.4} & \textbf{11.6} & \textbf{65.6} & \textbf{62.3} & \textbf{68.8} \\ 
\bottomrule[1.5pt]  
\end{tabular}
\end{table*}

\section{Experiments}

\subsection{Datasets and Metrics}
\subsubsection{Datasets} 
We conduct experiments on four popular and challenging video datasets: Ref-YouTube-VOS~\cite{URVOS}, Ref-DAVIS17~\cite{LREVOS}, A2D-Sentences~\cite{HuA2D}, JHMDB-Sentences~\cite{JHMDB}, \textcolor{black}{and MeViS~\cite{ding2023mevis}}. 
Ref-YouTube-VOS is a large-scale dataset that consists of 3,471 training videos with 12,913 expressions and 507 validation videos with 2,096 expressions.
Ref-DAVIS17 comprises 90 videos with 1,544 expressions, captured at 24 FPS. 
\textcolor{black}{A2D-Sentences is an actor and action segmentation dataset with 3,782 video sequences, 6,656 action descriptions, and 1 to 5 frame annotations per sequence.
JHMDB-Sentences includes 928 short video sequences and 928 descriptions covering 21 different action classes.}

\subsubsection{Metrics} 
We employ standard metrics for evaluation. All results are evaluated using the official evaluation servers or codes. 
For Ref-YouTube-VOS and Ref-DAVIS17, Region Jaccard $\mathcal{J}$, Boundary Accuracy $\mathcal{F}$, and their average $\mathcal{J} \& \mathcal{F}$~\cite{DAVIS} over all videos are reported. 
In addition to accuracy metrics, we evaluate the temporal consistency of segmentation on Ref-YouTube-VOS using our proposed MCS metric.
For A2D-Sentences and JHMDB-Sentences, we report Precision@K, mAP over 0.50:0.05:0.95, Overall IoU, and Mean IoU.
Precision@K measures the percentage of test samples with IoU scores exceeding thresholds K$\in$[0.5, 0.6, 0.7, 0.8, 0.9].

\vspace{-2mm}
\subsection{Implementation Details}
 
Following~\cite{sgmg,MTTR,ReferFormer}, we train our models on the Ref-YouTube-VOS train split and evaluate them on both Ref-YouTube-VOS and Ref-DAVIS17.
We also present results for our models pre-trained on static image datasets Ref-COCO/g/+~\cite{RefCOCO2,RefCOCO} and fine-tuned on Ref-YouTube-VOS.
For A2D-Sentences and JHMDB-Sentences, we fine-tune the pre-trained models on the A2D-Sentences train split and evaluate them on both A2D-Sentences and JHMDB-Sentences.

We maintain the same optimization strategies as~\cite{ReferFormer,sgmg} for fair comparison.
Our models are optimized with AdamW~\cite{AdamW} and a weight decay of 5$\times10^{-4}$.
The initial learning rates are set to 2.5$\times10^{-5}$ for the visual encoder and inter-frame collaboration module, 5$\times10^{-6}$ for the textual encoder, and 5$\times10^{-5}$ for the rest. 
The batch size is 4/2 for pre-training/main training and distributed on two 24GB RTX 3090 GPUs.
During pre-training, each sample consists of one static image, we train our models for 12 epochs with a learning rate decay by a factor of 10 at the 8th and 10th epochs.
During main training, each sample consists of 5 randomly sampled frames, we freeze the text encoder and train the models for 6/10 epochs with a learning rate decay by a factor of 10 at the 3/6th and 5/8th epochs, depending on whether pre-training was performed.
The loss function coefficients, $\lambda_{dice}$, $\lambda_{L1}$, $\lambda_{focal}$, $\lambda_{GIoU}$, $\lambda_{ce}$ are set to 5, 5, 2, 2, 1, respectively.
During inference, we follow~\cite{ReferFormer,sgmg} to perform video-level segmentation ($N$=frame number) on Ref-YouTube-VOS and per-clip segmentation ($N$=36) on Ref-DAVIS17. We selectively segment the top 25\% of ranked frames to construct memory for each video (see Table~\ref{tab:ablationmemframe} for the ablation study).

\begin{table*}[t!]
\setlength{\tabcolsep}{8pt}
\centering
\caption{Quantitative comparison to state-of-the-art R-VOS methods on A2D-Sentences} \label{tab:quantitativea2d}
\begin{tabular}{l | l | c | c c | c c c c c}
\toprule[1.5pt] 
\multirow{2}{*}{} & \multirow{2}{*}{} & \textbf{mAP} & \multicolumn{2}{c |}{\textbf{IoU}} & \multicolumn{5}{c}{\textbf{Precision}} \\ 
\midrule
Method & Backbone &  & Overall & Mean  & P@0.5 & P@0.6 & P@0.7 & P@0.8 & P@0.9 \\
\midrule

Hu \etal ~\cite{HuA2D} & VGG-16 & 13.2 &  47.4 & 35.0 & 34.8 & 23.6 & 13.3 & 3.3 & 0.1 \\
Gavrilyuk \etal ~\cite{GavrilyukA2D} & I3D & 19.8 & 53.6 & 42.1 & 47.5 & 34.7 & 21.1 & 8.0 & 0.2 \\
CMSA + CFSA ~\cite{CMSA} & ResNet-101 & - & 61.8 & 43.2 & 48.7 & 43.1 & 35.8 & 23.1 & 5.2 \\
ACAN~\cite{ACAN} & I3D & 27.4 & 60.1 & 49.0 & 55.7 & 45.9 & 31.9 & 16.0 & 2.0 \\
CMPC-V ~\cite{CMPC} & I3D & 40.4 & 65.3 & 57.3 & 65.5 & 59.2 & 50.6 & 34.2 & 9.8 \\
ClawCraneNet~\cite{Clawcranenet} & ResNet-50/101 & - & 63.1 & 59.9 & 70.4 & 67.7 & 61.7 & 48.9 & 17.1 \\
MTTR~\cite{MTTR} & VideoSwin-T & 46.1 & 72.0 & 64.0 & 75.4 & 71.2 & 63.8 & 48.5 & 16.9 \\
ReferFormer~\cite{ReferFormer} & VideoSwin-B & 55.0 & 78.6 & 70.3 & 83.1 & 80.4 & 74.1 & 57.9 & 21.2 \\
\textcolor{black}{VLT~\cite{VLT}} & \textcolor{black}{VideoSwin-B} & \textcolor{black}{55.6} & \textcolor{black}{78.8} & \textcolor{black}{70.5} & \textcolor{black}{83.1} & \textcolor{black}{80.7} & \textcolor{black}{74.5} & \textcolor{black}{58.5} & \textcolor{black}{21.7} \\
OnlineRefer~\cite{onlinerefer} & VideoSwin-B & - & 79.6 & 70.5 & 83.1 & 80.2 & 73.4 & 56.8 & 21.7 \\
HTML~\cite{HTML} & VideoSwin-B &  56.7 & 79.5 & 71.2 & 84.0 & 81.5 & 75.8 & 59.2 & 22.8 \\
\rowcolor[gray]{0.9} 
\textbf{\ourmethod{} (Ours)} & VideoSwin-B & \textbf{59.0} & \textbf{80.1} & \textbf{72.6} & \textbf{85.2} & \textbf{83.2} & \textbf{77.9} & \textbf{62.4} & \textbf{25.5} \\

\bottomrule[1.5pt] 
\end{tabular}
\end{table*}

\begin{table*}[t!]
\setlength{\tabcolsep}{8pt}
\centering
\vspace{-1mm}
\caption{Quantitative comparison to state-of-the-art R-VOS methods on JHMDB-Sentences} \label{tab:quantitativejhmdb}
\begin{tabular}{l | l | c | c c | c c c c c}
\toprule[1.5pt] 
\multirow{2}{*}{} & \multirow{2}{*}{} & \textbf{mAP} & \multicolumn{2}{c |}{\textbf{IoU}} & \multicolumn{5}{c}{\textbf{Precision}} \\ 
\midrule
Method & Backbone &  & Overall & Mean  & P@0.5 & P@0.6 & P@0.7 & P@0.8 & P@0.9 \\
\midrule

Hu \etal ~\cite{HuA2D} & VGG-16 & 17.8 &  54.6 & 52.8 & 63.3 & 35.0 & 8.5 & 0.2 & 0.0 \\
Gavrilyuk \etal ~\cite{GavrilyukA2D} & I3D & 23.3 & 54.1 & 54.2 & 69.9 & 46.0 & 17.3 & 1.4 & 0.0 \\
CMSA + CFSA ~\cite{CMSA} & ResNet-101 & - & 62.8 & 58.1 & 76.4 & 62.5 & 38.9 & 9.0 & 0.1 \\
ACAN~\cite{ACAN} & I3D & 28.9 & 57.6 & 58.4 & 75.6 & 56.4 & 28.7 & 3.4 & 0.0 \\
CMPC-V ~\cite{CMPC} & I3D & 34.2 & 61.6 & 61.7 & 81.3 & 65.7 & 37.1 & 7.0 & 0.0 \\
ClawCraneNet~\cite{Clawcranenet} & ResNet-50/101 & - & 64.4 & 65.6 & 88.0 & 79.6 & 56.6 & 14.7 & 0.2 \\
MTTR~\cite{MTTR} & VideoSwin-T & 39.2 & 70.1 & 69.8 & 93.9 & 85.2 & 61.6 & 16.6 & 0.1 \\
ReferFormer~\cite{ReferFormer} & VideoSwin-B & 43.7 & 73.0 & 71.8 & 96.2 & 90.2 & 70.2 & 21.0 & 0.3 \\
OnlineRefer~\cite{onlinerefer} & VideoSwin-B & - & 73.5 & 71.9 & 96.1 & 90.4 & 71.0 & 21.9 & 0.2 \\
\rowcolor[gray]{0.9} 
\textbf{\ourmethod{} (Ours)} & VideoSwin-B & \textbf{44.9} & \textbf{73.9} & \textbf{72.6} & \textbf{97.1} & \textbf{92.0} & \textbf{72.1} & \textbf{23.7} & \textbf{0.3} \\

\bottomrule[1.5pt] 
\end{tabular}
\vspace{-1mm}
\end{table*}

\begin{figure*}[t!]
\centering
\includegraphics[width=\textwidth]{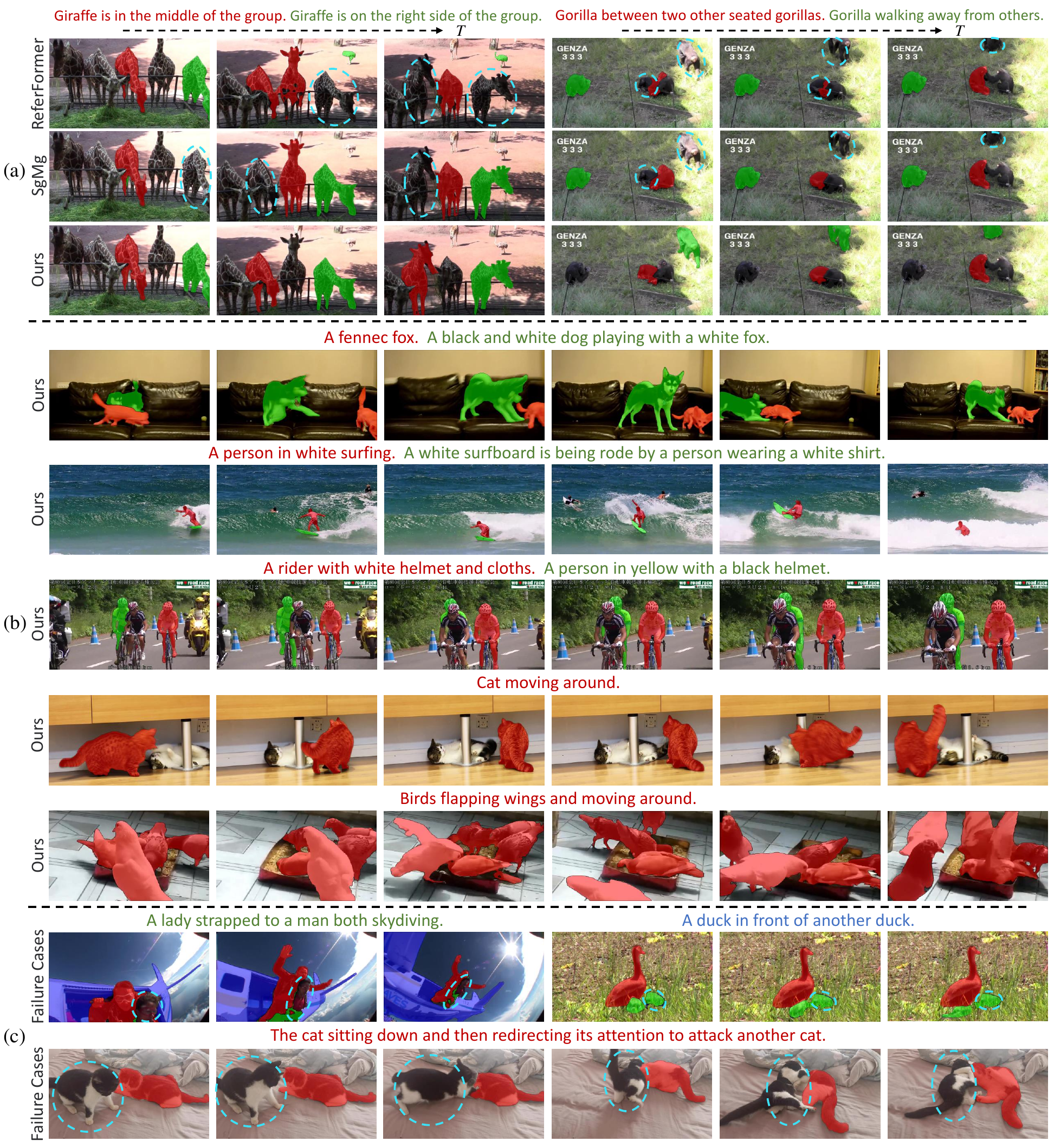}
\caption{\textcolor{black}{Qualitative results on Ref-YouTube-VOS and MeViS. (a) Our \ourmethod{} predicts temporally consistent results compared to ReferFormer~\cite{ReferFormer} and SgMg~\cite{sgmg}.
(b) \ourmethod{} can handle appearance and motion expressions in various challenging scenarios, including fast motion, objects with similar appearance, occlusion, and small objects.
(c) \ourmethod{} fails to improve mask quality without good reference masks.} 
}\label{fig:qualitative} 
\end{figure*}

\begin{figure*}[t!]
\centering
\includegraphics[width=\textwidth]{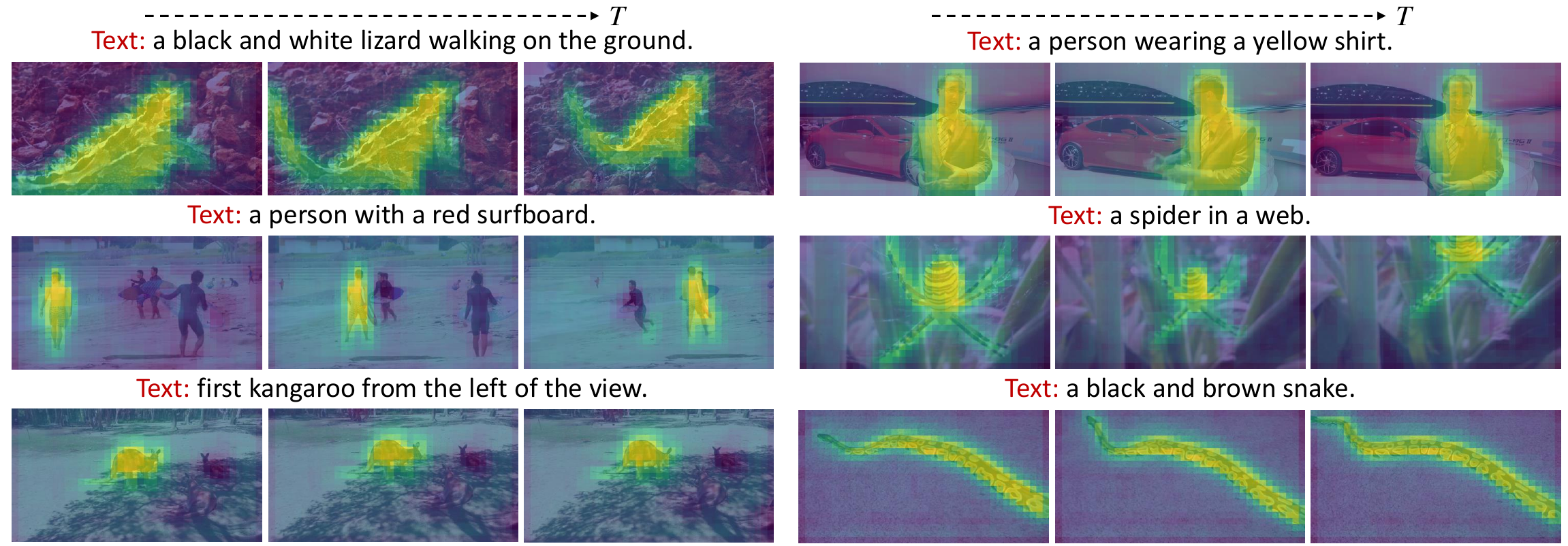}
\caption{Visualization of the affinity between the vision-mask joint representations of target frames and the global foreground representation in the hybrid memory.
Our feature aggregation extracts robust global representations to localize the correct targets.
}\label{fig:globalvisual} 
\end{figure*}

\subsection{Main Results}

\subsubsection{Results on Ref-YouTube-VOS}
Table~\ref{tab:mainresults} shows that our \ourmethod{} (with Swin-L) achieves the top-ranked performance of \textbf{67.1} $\mathcal{J} \& \mathcal{F}$ on Ref-YouTube-VOS, significantly outperforming the baseline ReferFormer~\cite{ReferFormer} by \textbf{4.7\% points}. 
Without pre-training on Ref-COCO/g/+, \ourmethod{} with VideoSwin-T achieves 59.8 $\mathcal{J} \& \mathcal{F}$ at 58 FPS, still surpassing ReferFormer while running about 1.2$\times$ faster. 
Despite its selective segmentation stage being built upon ReferFormer, \ourmethod{} exceeds all recent competitors, including OnlineRefer~\cite{onlinerefer}, HTML~\cite{HTML}, SgMg~\cite{sgmg}, and SOC~\cite{soc}, particularly in terms of temporal consistency.
This demonstrates the effectiveness of our approach.
Additionally, we integrate our hybrid memory with the recent work SgMg~\cite{sgmg} to create the end-to-end model S-\ourmethod{} for comparison. 
This integration further improves performance by a large margin.
Without pre-training on Ref-COCO/g/+, S-\ourmethod{} with VideoSwin-T achieves 61.5 $\mathcal{J} \& \mathcal{F}$, surpassing SgMg by 2.6\% points. 
These results further demonstrate our architecture is effective and generic.

\subsubsection{Results on Ref-DAVIS17} 
In Table~\ref{tab:mainresults}, 
\ourmethod{} shows outstanding generality on Ref-DAVIS17, remarkably surpassing the benchmark ReferFormer~\cite{ReferFormer} by \textbf{5.1\% points} in terms of $\mathcal{J}\&\mathcal{F}$. 
Specifically, \ourmethod{} with VideoSwin-T achieves 63.2 $\mathcal{J} \& \mathcal{F}$, while \ourmethod{} with Swin-L achieves the top-ranked performance at \textbf{65.6} $\mathcal{J} \& \mathcal{F}$.

\subsubsection{Temporal Consistency}
In Table~\ref{tab:mainresults}, we compare \ourmethod{} with other open-sourced R-VOS models on Ref-YouTube-VOS using the proposed MCS metric, which evaluates the temporal consistency of segmentation across thresholds from 0.1 to 0.9. 
\ourmethod{} significantly outperforms ReferFormer~\cite{ReferFormer}, MTTR~\cite{MTTR}, OnlineRefer~\cite{onlinerefer}, SgMg~\cite{sgmg}, and SOC~\cite{soc} across all thresholds due to the temporal consistency modeling.
When Swin-L is used as the visual encoder, \ourmethod{} outperforms the baseline ReferFormer by 20.3\%, 21.7\%, and 52.6\% under MCS@0.1, MCS@0.5, MCS@0.9, respectively.
Although \ourmethod{} with VideoSwin-T employs a smaller backbone compared to ReferFormer with Swin-L, it still achieves notably higher MCS.
These results prove the effectiveness of \ourmethod{} in improving temporal consistency.
Moreover, the MCS decreases remarkably for thresholds above 0.5, indicating the future research direction of developing R-VOS models that can produce both temporally consistent and accurate segmentation outcomes.

\subsubsection{Results on A2D-Sentences \& JHMDB-Sentences} 
In Table~\ref{tab:quantitativea2d} and  Table~\ref{tab:quantitativejhmdb}, we further evaluate our models on the A2D-Sentences and JHMDB-Sentences, comparing them with other state-of-the-art methods. 
As only a few frames in each video sequence are annotated for training, the models are less impacted by temporally consistent modeling on these two datasets.
Despite this, our models exceed other R-VOS methods in all metrics, further demonstrating the effectiveness of our approach.

\subsubsection{Inference Speed and Parameters}
We perform selective conditional segmentation and efficient inter-frame collaboration for temporally consistent R-VOS.
\textcolor{black}{By conducting conditional segmentation \emph{only} on top-ranked (reference) frames, \ie, avoiding intensive cross-modal FPN decoding on remaining frames, HTR with VideoSwin-T achieves 58 FPS on a single RTX 3090 GPU. This is faster than ReferFormer~\cite{ReferFormer} (50 FPS) and comparable to the recent SOC~\cite{soc} (59 FPS) using the same backbone. Additionally, HTR with VideoSwin-B and Swin-L runs at 36 and 42 FPS, respectively. Despite having approximately 12\% (22M) more parameters due to the hybrid memory, our model runs faster and outperforms ReferFormer by up to 5\% points on Ref-YouTube-VOS, demonstrating good performance trade-off.}

\begin{table}[t!]
\centering
\footnotesize
\vspace{-2mm}
\caption{\textcolor{black}{Quantitative comparison on MeViS}} \label{tab:mevis} 
\textcolor{black}{
\begin{tabular}{l c c c}
\toprule[1.5pt] 
Methods & \( \mathcal{J} \)\&\( \mathcal{F} \) & $\mathcal{J}$ & $\mathcal{F}$ \\ 
\midrule 
URVOS~\cite{URVOS} & 27.8 & 25.7 & 29.9 \\
LBDT~\cite{LBDT} & 29.3 & 27.8 & 30.8 \\
MTTR~\cite{MTTR} & 30.0 & 28.8 & 31.2 \\
ReferFormer~\cite{ReferFormer} & 31.0 & 29.8 & 32.2 \\
VLT+TC~\cite{ding2023mevis} & 35.5 & 33.6 & 37.3 \\
LMPM~\cite{ding2023mevis} & 37.2 & 34.2 & 40.2 \\
\rowcolor[gray]{0.9} 
\textbf{\ourmethod{} (Ours)} & \textbf{42.7} & \textbf{39.9} & \textbf{45.5} \\
\bottomrule[1.5pt] 
\end{tabular}
}
\vspace{-2mm}
\end{table}

\subsubsection{\textcolor{black}{Results on MeViS}}
\textcolor{black}{In Table~\ref{tab:mevis}, we further evaluate \ourmethod{} with the tiny version backbone on the recent MeViS~\cite{ding2023mevis} dataset. Training from scratch on the MeViS train split, our model achieves 42.7 \( \mathcal{J} \)\&\( \mathcal{F} \), outperforming ReferFormer by \textbf{11.7}\% points and demonstrating advanced performance. This improvement is more significant because the MeViS dataset focuses on moving objects and longer videos, challenging previous models with temporal consistency issues.}

\subsection{Qualitative Results}

We visualize some typical segmentation results in Fig.~\ref{fig:qualitative}.
ReferFormer and SgMg face challenges when dealing with temporal context changes, such as the target giraffe is not always in the middle, and the existence of other similar-looking objects like giraffes and gorillas.
In contrast, \ourmethod{} predicts correct and temporally consistent object masks, thanks to its temporal consistency modeling.
\textcolor{black}{Additionally, \ourmethod{} can handle appearance and motion expressions in various challenging scenarios, including fast motion, occlusion, objects with similar appearance, and small objects.}

\subsection{Limitations}
Our paradigm relies upon the availability of a few high-quality reference masks. 
However, the selective segmentation stage may not provide any good reference mask for some challenging videos. \textcolor{black}{Two such failure cases are shown in Fig.~\ref{fig:qualitative}(c).} In such cases, our paradigm will obviously be unable to improve the segmentation quality. 
Nevertheless, our approach still significantly outperforms existing methods without using any ground-truth reference mask.
Future research could develop more effective selective referring segmentation models.

\subsection{Visualization of Global Token Association}

In Fig.~\ref{fig:globalvisual}, we visualize the affinity maps $\hat{X}^{q}$ between the vision-mask joint representations $F^{J} \in{\mathbb{R} ^{C \times HW}}$ of target frames and the global foreground representation $E^{m} \in{\mathbb{R} ^{C \times 1}}$ in the hybrid memory (as defined in Eq.~\ref{equ:globalmem_sim}).
Our global tokens successfully localize the referred objects thanks to the robust global token aggregation.

\begin{table}[t!]
\centering
\setlength{\tabcolsep}{4pt}
\caption{Comparison with well-trained VOS models on Ref-YouTube-VOS. E.: end-to-end. Add. Enc.: additional visual encoder. Add. Data: additional training data. X-\ourmethod{}: replace our \ourmemory{} with XMem~\cite{XMem}. \ourmethod{} is end-to-end and achieves superior performance under fair comparison
} \label{tab:ablationexternalvos}
\begin{tabular}{l c c c c c c}
\toprule[1.5pt] 
Methods & E. & Add. Enc. & Add. Data & \( \mathcal{J} \)\&\( \mathcal{F} \) & Prop. FPS \\
\midrule 
ReferFormer & \checkmark & $\times$ & $\times$ & 62.4 & - \\
ReferFormer + CFBI & $\times$ & \checkmark & \checkmark & 63.3 & 7 \\
ReferFormer + STCN & $\times$ & \checkmark & \checkmark & 63.6 & 69 \\
ReferFormer + XMem & $\times$ & \checkmark & \checkmark & 64.4 & 61 \\ 
X-\ourmethod{} (Ours) & \checkmark & $\times$ & $\times$ & \textcolor{black}{65.0} & \textcolor{black}{168} \\ 
\ourmethod{} (Ours) & \checkmark & $\times$ & $\times$ & \textbf{67.1} & \textbf{181} \\ 
\bottomrule[1.5pt] 
\end{tabular}
\end{table}

\begin{table}[t!]
\centering
\caption{Ablation of memory components. Collaborative local and global components achieve the best performance} \label{tab:ablationmemory} 
\begin{tabular}{l c c c c}
\toprule[1.5pt] 
Methods & \( \mathcal{J} \)\&\( \mathcal{F} \) & MCS@0.1 & MCS@0.5 \\ 
\midrule 
w/o memory & 59.4 & 50.5 &40.8\\
w/ Local & 62.5 & 60.8 &50.7\\
w/ Global & 62.9 & \textbf{63.3} &53.0\\
w/ Hybrid & \textbf{64.2} & \textbf{63.3} &	\textbf{53.8}\\
\bottomrule[1.5pt] 
\end{tabular}
\vspace{-1mm}
\end{table}

\begin{table*}[t!]
\centering
\begin{minipage}{.3\textwidth}
\centering
\caption{Ablation of memorized ratio} \label{tab:ablationmemframe}
\setlength{\tabcolsep}{6pt}
\begin{tabular}{l c c c}
\toprule[1.5pt] 
Ratio ($k$) & \( \mathcal{J} \)\&\( \mathcal{F} \) & \( \mathcal{J} \) & \( \mathcal{F} \) \\ 
\midrule 
1\% & 63.5 & 62.0 & 64.9 \\
25\% & \textbf{64.2} & \textbf{62.7} & \textbf{65.7} \\
50\% & 63.9 & 62.5 & 65.3 \\
75\% & 63.2 & 61.7 & 64.6 \\
\bottomrule[1.5pt] 
\end{tabular}
\end{minipage}
\hfill
\begin{minipage}{.3\textwidth}
\centering
\caption{Ablation of strides} \label{tab:match_res}
\setlength{\tabcolsep}{6pt}
\begin{tabular}{l c c c}
\toprule[1.5pt] 
Stride & \( \mathcal{J} \)\&\( \mathcal{F} \) & $\mathcal{J}$ & $\mathcal{F}$\\ 
\midrule 
8 & 63.4 & 62.1 & 64.8 \\
16 & \textbf{64.2} & \textbf{62.7} & \textbf{65.7}\\
32 & 63.3 & 61.8 & 64.8 \\
\bottomrule[1.5pt] 
\end{tabular}
\end{minipage}
\hfill
\begin{minipage}{.35\textwidth}
\centering
\caption{\textcolor{black}{Ablation of
number of training frames}}\label{tab:frame_sample}
\setlength{\tabcolsep}{6pt}
\textcolor{black}{
\begin{tabular}{l c c c c}
\toprule[1.5pt] 
$N$ & MCS@0.5 & \( \mathcal{J} \)\&\( \mathcal{F} \) & $\mathcal{J}$ & $\mathcal{F}$\\ 
\midrule 
3 & 51.0 & 62.0 & 60.6 & 63.4 \\
5 & 53.8 & 64.2 & 62.7 & 65.7\\
10 & 54.6 & 64.9 & 63.4 & 66.4 \\
\bottomrule[1.5pt] 
\end{tabular}
}
\end{minipage}
\end{table*}

\begin{table}[t!]
\centering
\caption{\textcolor{black}{Performance on videos with different numbers of frames}}\label{tab:improve_num_frame}
\textcolor{black}{
\begin{tabular}{l c c c}
\toprule[1.5pt] 
Method & $0 < N \leq 50$ & $50 < N \leq 80$ & $N > 80$ \\ 
\midrule 
ReferFormer & 59.8 & 59.4 & 61.5 \\
HTR (Ours) & \textbf{66.9} & \textbf{63.1} & \textbf{67.0} \\ 
\bottomrule[1.5pt] 
\end{tabular}
}
\vspace{-1mm}
\end{table}

\subsection{Ablation Study}

\subsubsection{Well-trained VOS models}
In Table~\ref{tab:ablationexternalvos}, we compare \ourmethod{} with the combination of the benchmark ReferFormer~\cite{ReferFormer} and well-trained temporal propagation models, \ie, CFBI~\cite{cfbi}, STCN~\cite{STCN}, and XMem~\cite{XMem}. 
Following ReferFormer, we use Swin-L as the visual encoder in this experiment (VideoSwin-T is used in subsequent ablations).
We perform comparisons using 360p frames and adopt high-score frames as reference frames for fair comparison.
Our end-to-end \ourmethod{} significantly surpasses others by over 2.7\% and runs up to 26$\times$ faster.
Moreover, replacing the \ourmemory{} with XMem in our paradigm results in a 2.1\% lower \( \mathcal{J} \)\&\( \mathcal{F} \).
XMem designs sensory memory to propagate the ground-truth mask of the first frame which does not align with our challenge of propagating imperfect reference masks from any frames. In contrast, \ourmethod{} extracts robust global context and improves matching and memory extracted from imperfect reference masks. 
These results confirm the effectiveness of our hybrid memory and paradigm.

\subsubsection{Hybrid Memory}
In Table~\ref{tab:ablationmemory}, we evaluate the individual contributions of the components within our \ourmemory{}. Both local and global components exhibit strong temporal propagation ability in our paradigm. Notably, global tokens outperform local memory.
Our global tokens employ weighted feature aggregation to suppress noise and form robust global representations. 
This is crucial for improving the temporal consistency and mask quality during propagation especially when the initial reference masks are automatically generated and imperfect.
Overall, the collaboration of local and global components leads to superior performance.

\subsubsection{Memorized Frames}
In Table~\ref{tab:ablationmemframe}, we evaluate the performance of \ourmethod{} across different ratios of memorized frames.
\ourmethod{} dynamically selects the top $k$ percent of ranked frames (frame numbers are rounded up) to construct the \ourmemory{}.
The results show that memorizing the top 25\% of ranked frames to segment the remaining 75\% achieves the best performance.

\subsubsection{Different Feature Strides}
In this work, \ourmethod{} performs inter-frame collaboration using features with stride 16, a common setting for memory modules~\cite{STM,XMem}.
Table~\ref{tab:match_res} presents our evaluation of different strides, demonstrating that the common setting achieves the best performance.

\subsubsection{\textcolor{black}{Number of Training Frames}}
\textcolor{black}{We follow previous works such as~\cite{ReferFormer,sgmg,R2VOS} to randomly sample 5 frames during training for fair comparison. This \emph{random} sampling enables the model to learn both long-term and short-term matching for temporally consistent R-VOS. In Table~\ref{tab:frame_sample}, we also explore different numbers of training frames. Our model consistently performs well, with $N$=5 demonstrating the best trade-off.}

\subsubsection{\textcolor{black}{Videos with Different Numbers of Frames}}
\textcolor{black}{In Table~\ref{tab:improve_num_frame}, we compare the performance (\( \mathcal{J} \)\&\( \mathcal{F} \)) of our model and the baseline ReferFormer on Ref-DAVIS17 videos with varying frame numbers. \ourmethod{} achieves significantly better results than ReferFormer across both short and long videos.}

\section{Conclusion}

This paper proposes a novel end-to-end paradigm for referring video object segmentation (R-VOS) that effectively addresses the temporal inconsistency issue.
Unlike previous solutions, our paradigm associates and propagates spatio-temporal features in an end-to-end fashion for achieving temporally consistent segmentation.
To handle spatio-temporal propagation in the presence of noise, we introduced the \ourmemory{} with inter-frame collaboration to conduct robust multi-granularity association.
Furthermore, we proposed a new Mask Consistency Score (MCS) metric to evaluate the temporal consistency of video segmentation. 
Our results demonstrate that incorporating explicit temporal instance consistency modeling not only effectively improves temporal consistency, but also greatly enhances segmentation quality, achieving top-ranked performance on popular R-VOS benchmarks.
We hope our paradigm will also benefit other video segmentation tasks in the future.

\bibliography{reference}

%
\IEEEpeerreviewmaketitle

\IEEEpeerreviewmaketitle

\end{document}